\documentclass[11pt,a4paper]{article}
\usepackage[hyperref]{eacl2021}
\usepackage{times}
\usepackage{latexsym}
\usepackage{url}
\usepackage{graphicx}

\usepackage{enumitem}
\usepackage{multirow}
\usepackage{array}

\makeatletter
\newcommand\footnoteref[1]{\protected@xdef\@thefnmark{\ref{#1}}\@footnotemark}
\makeatother

\usepackage{microtype}

\aclfinalcopy % Uncomment this line for the final submission
\title{A Systematic Review of\\Reproducibility Research in Natural Language Processing}

\author{Anya Belz \\
  University of Brighton, UK \\
  \texttt{a.s.belz@brighton.ac.uk} \And
  Shubham Agarwal \\
  Heriot Watt University, UK \\
  \texttt{sa201@hw.ac.uk} \AND
  Anastasia Shimorina \\
  Universit\'e de Lorraine / LORIA, France \\
  \texttt{anastasia.shimorina@loria.fr} \And
  Ehud Reiter \\
  University of Aberdeen, UK \\
  \texttt{e.reiter@abdn.ac.uk}}

\date{}

\begin{document}
\maketitle
\begin{abstract}
%\vspace{-.1cm}
Against the background of what has been termed a reproducibility crisis in science, the NLP field is becoming increasingly interested in, and conscientious about, the reproducibility of its results. The past few years have seen an impressive range of new initiatives, events and active research in the area. However, the field is far from reaching a consensus about how reproducibility should be defined, measured and addressed, with diversity of views currently increasing rather than converging. With this focused contribution, we aim to provide a wide-angle, and as near as possible complete, snapshot of current work on reproducibility in NLP, delineating differences and similarities, and providing 
pointers to common denominators.
%\vspace{-.05cm}
\end{abstract}

\section{Introduction}\label{sec:intro}

%\vspace{-.1cm}
Reproducibility is one of the cornerstones of scientific research: inability to reproduce results is, with few exceptions, seen as casting doubt on their validity. Yet it is surprisingly hard to achieve, 70\% of scientists reporting failure to reproduce someone else's results, and more than half reporting failure to reproduce their own, a state of affairs that has been termed the `reproducibility crisis' in science   \cite{baker2016reproducibility}. Following a history of troubling evidence regarding difficulties in reproducing results \cite{pedersen2008empiricism,mieskes-etal-2019-community}, where 24.9\% of attempts to reproduce own results, and 56.7\% of attempts to reproduce another team's results, are reported to fail to reach the same conclusions \cite{mieskes-etal-2019-community}, the machine learning (ML) and natural language processing (NLP) fields have recently made great strides towards recognising the importance of, and addressing the challenges posed by, reproducibility: there have been several workshops on reproducibility in ML/NLP including the Reproducibility in ML Workshop at ICML'17, ICML'18 and ICLR'19; the Reproducibility Challenge at ICLR'18, ICLR'19, 
NeurIPS'19,
and NeurIPS'20;
LREC'20 had a reproducibility track and shared task \cite{branco-etal-2020-shared}; and NeurIPS'19 had a  reproducibility programme comprising a code submission policy, a reproducibility challenge for ML results, and the ML Reproducibility checklist \cite{whitaker2017}, later also adopted by EMNLP'20 and AAAI'21. Other conferences have foregrounded reproducibility via calls, chairs' blogs,\footnote{\scriptsize\url{https://2020.emnlp.org/blog/2020-05-20-reproducibility}} special themes and social media posts. Sharing code, data and supplementary material providing details about data, systems and training regimes\footnote{There are some situations where it is difficult to share data, e.g.\ because the data is commercially confidential or because it contains sensitive personal information.   But the increasing expectation in NLP is that authors should share as much as possible, and justify cases where it is not possible.} is firmly established in the ML/NLP community, virtually all main events now encouraging and making space for it.
Reproducibility even seems set to become a standard part of reviewing processes via checklists.
Far from beginning to converge in terms of standards, terminology and underlying definitions, however, this flurry of work is currently characterised by growing diversity in all these respects. We start below by surveying concepts and definitions in reproducibility research, areas of particular disagreement, 
and identify categories of work in current NLP reproducibility research. We then use the latter to structure the remainder of the paper.

%\vspace{-.15cm}
\paragraph{Selection of Papers:}
We conducted a structured review employing a stated systematic process for identifying all papers in the field that met specific criteria.  Structured reviews are a type of meta-review  more common in fields like medicine but beginning to be used more in NLP \citep{reitercl18,howcroft-etal-2020-twenty}.

Specifically, we selected papers as follows. We searched the ACL Anthology for titles containing either of the character strings \textit{reproduc} or \textit{replica}, either capitalised or not.\footnote{{\texttt{grep -E `title *=.*(r$|$R)$\backslash$\}*(eproduc$|$epl ica)' anthology.bib}}} This yielded 47 papers; following 
inspection we excluded 12 of the papers as not being about reproducibility in the present sense.\footnote{Most were about annotation and data replication.} 
We found 25 additional papers in non-ACL NLP/ML sources,
and a further 7 in other fields. 

\begin{figure}[t]
\vspace{-.4cm}
    \hspace{-.8cm}\hspace{-.2cm}\includegraphics[width=.59\textwidth]{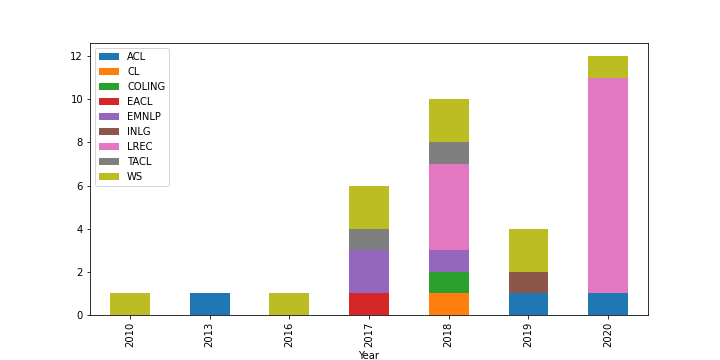}
\vspace{-.6cm}
    \caption{35 papers from ACL Anthology search, by year and venue.\footnoteref{code}}
    \label{fig:repr-graph}
%\vspace{-.2cm}
\end{figure}

Figure~\ref{fig:repr-graph} shows\footnote{\label{code}Data and code: \url{https://github.com/shubhamagarwal92/eacl-reproducibility}} how the 35 papers from the ACL Anthology search are distributed over years: one paper a year at most until 2017/18 when interest seems to have increased spontaneously, before dropping off again. 
The renewed high numbers 
for 2020 are almost entirely due to the LREC REPROLANG shared task (see Section~\ref{sec:surveys-multi} below).

\section{Terminology and Frameworks}\label{sec:terminology}

Reproducibility research in NLP and beyond uses a bewildering range of closely related terms, often with conflicting meaning, including reproducibility, repeatability, replicability, recreation, re-run, robustness, repetition and generalisability.
The fact that no formal definition of any of these terms singly, let alone in relation to each other, is generally accepted as standard, or even predominant, in NLP at present, is clearly a problem for a survey paper. 
In this section, we review usage
before identifying common-ground terminology that will enable us to talk about the research we survey.

The two most frequently used `R-terms', \textit{reproducibility} and \textit{replicability}, are also the most problematic. For the ACM \cite{acm2020artifact}, results have been \textit{reproduced} if ``obtained in a subsequent study by a person or team other than the authors, using, in part, artifacts provided by the author," and \textit{replicated} if ``obtained in a subsequent study by a person or team other than the authors, without the use of author-supplied artifacts" (although initially the terms were defined the other way around\footnote{ACM swapped definitions of the two terms when prompted by NISO to ``harmonize its terminology and definitions with those used in the broader scientific research community." \cite{acm2020artifact}.}). 
The definitions are tied to team and software (artifacts), but it is unclear how much of the latter have to be the same for reproducibility, and how different the team needs to be for either concept. 

\citet{rougier2017sustainable} tie definitions (just) to new vs.\ original software: ``\textit{Reproducing} the result of a computation means running the same software on the same input data and obtaining the same results. [...] \textit{Replicating} a published result means writing and then running new software based on the description of a computational model or method provided in the original publication, and obtaining results that are similar enough to be considered equivalent.'' It is clear from the many reports of failures to obtain `same results' with `same software and data' in recent years that the above definitions raise practical questions such as how to tell `same software' from `new software,' and how to determine equivalence of results.

\citet{wieling-etal-2018-squib} define \textit{reproducibility} as ``the exact re-creation of the results reported in a publication using the same data and methods," but then discuss (failing to) \textit{replicate} results without defining that term, while also referring to the ``unfortunate swap'' of the definitions of the two terms put forward by \citet{drummond2009replicability}. 

\citet{whitaker2017}, followed by \citet{schloss2018identifying}, tie definitions to data as well as code:

\vspace{.3cm}
\begin{small}
\begin{tabular}{|c|c|c|c|}
\hline
\multicolumn{1}{|c}{} & & \multicolumn{2}{c|}{\textbf{Data}} \\
\cline{3-4}
\multicolumn{1}{|c}{} & & \textit{Same} & \textit{Different} \\
\hline
\multirow{2}{*}{\textbf{Code}} & \textit{Same} & Reproducible & Replicable \\
\cline{2-4}
    & \textit{Different} & Robust & Generalisable \\
\hline
\end{tabular}   
\end{small}
\vspace{.25cm}

\noindent The different definitions of \textit{reproducibility} and \textit{replicability} above, put forward in six different contexts, are not compatible with each other. Grappling with degrees of similarity between properties of experiments such as the team, data and software involved, and between results obtained, each draws the lines between terms differently, and moreover variously treats reproducibility and replicability as properties of either systems or results. All are patchy, not accounting for some circumstances, e.g.\ a team reproducing its own results, not defining some concepts, e.g.\ sameness, or not specifying what can serve as a `result,' e.g.\ leaving the status of human evaluations and dataset recreations unclear.

The extreme precision of the definitions of the International Vocabulary of Metrology (VIM) \cite{jcgm2012international} (which the ACM definitions are supposed to be based on but aren't quite) offers a common terminological denominator. The VIM definitions of \textit{reproducibility} and \textit{repeatability} (no other R-terms are defined) are entirely general, made possible by two key differences compared to the NLP/ML definitions above. Firstly, in a key conceptual shift,
reproducibility and repeatability are properties of \textbf{measurements} (not of systems or abstract findings). The important difference  is that the concept of reproducibility now references a specified way of obtaining a \textbf{measured quantity value} (which can be an evaluation metric, statistical measure, human evaluation method, etc.\ in NLP). Secondly, reproducibility and repeatability are defined as the \textit{precision} of a measurement under specified conditions, i.e.\ the distribution of the quantity values obtained in \textit{repeat} (or \textit{replicate}) measurements.

In VIM, \textbf{repeatability} is the precision of measurements of the same or similar object obtained under the same conditions, as captured by a specified set of \textbf{repeatability conditions}, whereas \textbf{reproducibility} is the precision of measurements of the same or similar object obtained under different conditions, as captured by a specified set of \textbf{reproducibility conditions}. See Appendix~\ref{sec:appendix-a} for a full set of VIM definitions of the bold terms above. 

To make the VIM terms more recognisable in an NLP context, we also call repeatability \textbf{reproducibility under same conditions}, and (VIM) reproducibility \textbf{reproducibilty under varied conditions}. Finally, we refer to experiments carrying out repeat measurements regardless of same/varied conditions as `reproduction studies.'

\vspace{.1cm}
    
\paragraph{Categories of Reproducibility Research:}

Using the definitions above, the work we review in the remainder of the paper falls into three categories (corresponding to Sections~\ref{sec:reproduction-attempts}--\ref{sec:surveys-multi}):

\textbf{\textit{Reproduction under same conditions:}}
As near as possible exact {recreation} or reuse of an existing system and evaluation set-up, and comparison of results.\footnote{This excludes the countless cases where results for a previous method are used as a baseline or other comparitor, but experiments are not run again.} 

\textbf{\textit{Reproduction under varied conditions:}}
Reproduction studies with deliberate variation of one or more aspects of system and/or measurement, and comparison of results. 

\textbf{\textit{Multi-test studies:}} Multiple reproduction studies connected e.g.\ because of an overall multi-test design, and/or use of same methodology.

\noindent 

\begin{figure*}[th]
    \centering
\includegraphics[width=.875\textwidth]{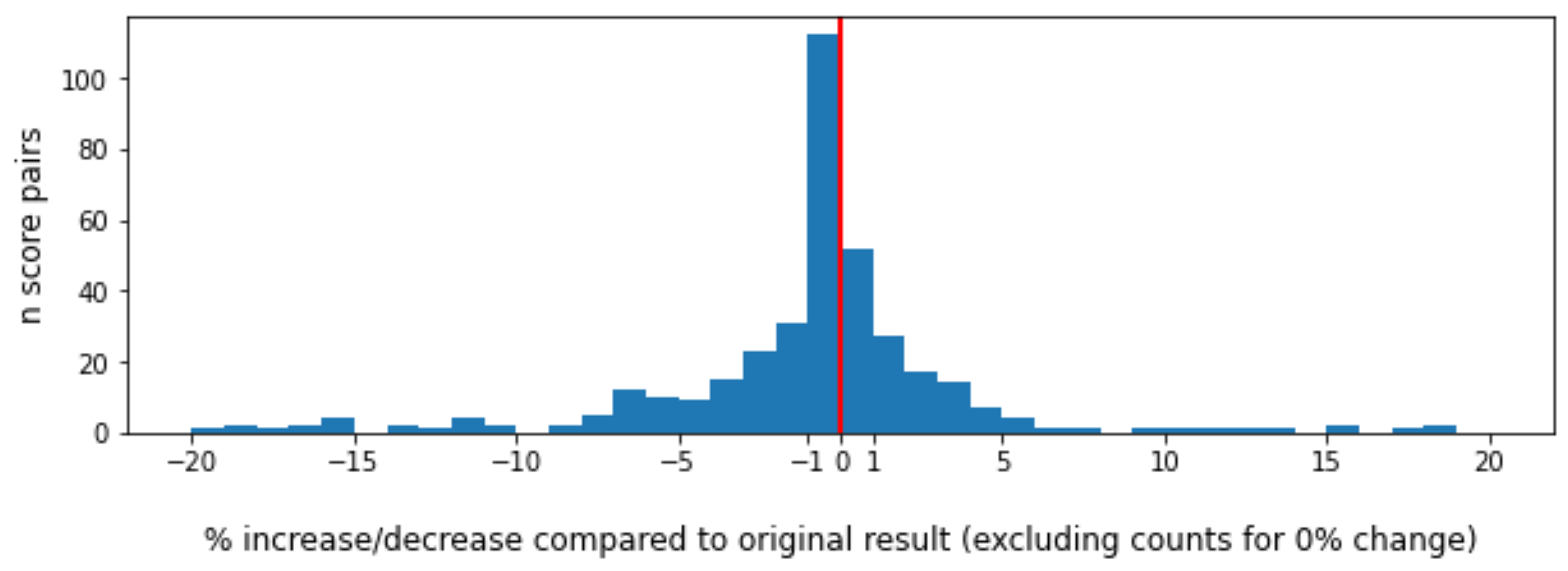} \\
    \caption{Histogram of percentage differences between original and reproduction scores (bin width = 1; clipped to range -20..20).}
    \label{fig:percentage-differences}
\end{figure*}

\section{Reproduction Under Same Conditions}\label{sec:systems}\label{sec:reproduction-attempts}

Papers reporting reproductions under same conditions account for the bulk of NLP reproducibility research to date. 
The difficulty of achieving `sameness of system' has taken up a lot of the discussion space. As stressed by many papers \cite{crane-2018-questionable,millour-etal-2020-repliquer}, recreation attempts have to have access to code, data, full details/assumptions of algorithms,
parameter settings, software and dataset versions, initialisation details, random seeds, run-time environment, hardware specifications, etc.  

A related and striking finding, confirmed by multiple repeatability studies, is that results often depend in surprising ways and to surprising degrees on seemingly small differences in model parameters and settings, such as rare-word thresholds, treatment of ties, or case normalisation \cite{fokkens-etal-2013-offspring,van2013reusable,dakota-kubler-2017-towards}.  
Effects are often discovered during system recreation from incomplete information, when a range of values is tested for missing details. 
The concern is that the ease with which such NLP results are perturbed casts doubt on their generalisability and robustness.

The difficulties in recreating, or even just re-running, systems with same results have led to growing numbers of reproducibility checklists \cite{olorisade2017reproducibility,pineau2020checklist}, and 
tips for making system recreation easier, e.g.\ the PyTorch \citep{paszke2017automatic} recommended settings.\footnote{\url{https://pytorch.org/docs/stable/notes/randomness.html}}

We analysed reproduction studies under same conditions from 34 pairs of papers, and identified 549 individual score pairs where reproduction object, method and outcome were clear enough to include in comparisons (for a small number of papers this meant excluding some scores).
Table~\ref{tab:repro-attempts} in Appendix~\ref{app:table-same} provides a summary of the results. 
In 36 cases, the reproduction study did not produce scores, e.g.\ because resource limits were reached, or code didn't work. This left 513 cases where the reproduction study produced a value that could be compared to the original score. 
Out of these, just 77 (14.03\%) were exactly the same. Out of the remaining 436 score pairs, in 178 cases (40.8\%), the reproduction score was better than the original, and in 258 cases (59.2\%) it was worse.

We also examined the size of the differences between original and reproduction scores. For this purpose we computed percentage change (increase or decrease) for each score pair, and looked at the distribution of size and direction of change. For this analysis, we excluded score pairs where one or both scores were 0, as well as 4 very large outliers (all increases). Results are shown in the form of a histogram with bin width 1 (and clipped to range -20..20) in Figure~\ref{fig:percentage-differences}. The plot clearly shows the imbalance between worse (60\% of non-same scores) and better (40\%) reproduction scores. The figure also shows that a large number of differences fall in the -1..1 range. 
However, the majority of differences, or 3/5, are greater than +/-1\%, and about 1/4 are greater than +/-5\%.

\section{\hspace{-.15cm}Reproduction Under Varied Conditions}\label{sec:conf-finding}

Reproduction studies under varied conditions 
deliberately vary one or more aspects of system, data or evaluation in order to explore if similar results can be obtained. There are far fewer papers of this type (see Table~\ref{tab:findings-attempts} in Appendix~\ref{app:results-varied} for an overview) than  papers reporting reproduction studies under same conditions; however, note that we are not including papers here that use an existing method for a new language, dataset or domain, without controlling for other conditions being the same in experiments. 

\citet{horsmann-zesch-2017-lstms} pick up strong results by \citet{plank-etal-2016-multilingual} showing LSTM tagging to outperform CRF and HMM taggers, and test whether they can be confirmed for datasets with finer-grained tag sets.  
Using 27 corpora (21 languages) with finer-grained tag sets,  
they systematically compare results for the 3 models,
and show that LSTMs do perform better, and that 
their superiority grows in proportion to tag set size. 

\citet{htut-etal-2018-grammar} recreate the grammar induction model PRPN \cite{shen2018neural}, testing different versions 
with different data.
PRPN is confirmed to be ``strikingly effective" at latent tree learning. 
In a subsequent repeat study under same conditions, \citet{htut-etal-2018-grammar-induction} test PRPN using the authors' own code, obtaining the same headline result.

\citet{millour-etal-2020-repliquer} attempt to get the POS tagger for Alsatian from \citet{magistry2018etiquetage} to work with the same accuracy for a different dataset.  
Collaborating with, and using resources provided by, the original authors and recreating some where necessary, the best result obtained was 0.87 compared to the original 0.91.

\citet{abdellatif-elgammal-2020-ulmfit} varied  conditions of reproduction for classification results by \citet{howard2018universal}, 
and were able to confirm outcomes for three new non-English datasets, in all three respects (value, finding, conclusion) identified by \citet{cohen2018three}. 

\citet{plucinski-etal-2020-closer} and \citet{garneau-etal-2020-robust} both find that the cross-lingual word embedding mappings proposed by \citet{artetxe-etal-2018-robust} yield worse results on more distant language pairs.

\citet{vajjala-rama-2018-experiments}'s automatic essay scoring classification system was tested on different datasets and/or languages in three studies \cite{arhiliuc-etal-2020-language, caines-buttery-2020-reprolang, huber-coltekin-2020-reproduction} all of which found performance to drop on the new data.

\section{Multi-test and Multi-lab Studies}\label{sec:surveys-multi}

Work in this category is \textit{multi-test}, in the sense of involving multiple reproduction studies, 
in a uniform framework using uniform methods. Some of it is also \textit{multi-lab} in that reproduction studies are carried out by more than one research team. For example, in one multi-test repeatability study, \citet{wieling-etal-2018-squib} 
randomly select five papers each from ACL'11 and ACL'16 for which code/data was available.
In a uniform design, original authors were contacted for help if needed, a maximum time limit of 8h was imposed, and all work was done by the same Masters student. 
It's not clear how scores were selected (not all are attempted), and reasons for failure are not always clear even from linked material. Of the 120 score pairs obtained, 60 were the same, 12 reproduction scores were better, 22 were worse, and 26 runs failed (including exceeding the time limit). See Table~\ref{tab:repro-attempts} for summary.

\citet{antonio-rodrigues-etal-2020-reproduction} recreated six SemEval'18 systems from the Argument Reasoning Comprehension Task, following system descriptions and/or reusing code, with no time limit. Scores were the same for one system, and within +/- 0.036 points for the other five; the SemEval ranking was exactly the same. Systems were also run on a corrected version of the shared-task data (which contained unwitting clues). This resulted in much lower scores casting doubt on the validity of the original shared task results.

REPROLANG \cite{branco-etal-2020-shared} is so far the only multi-lab (as well as multi-test) study of reproducibility in NLP. It was run as a selective shared task, and required participants to conform to uniform rules. 11 papers were selected for reproduction via an open call and direct selection. Participants had to `reproduce the paper,' 
using information contained/linked in it.
Participants submitted (a) a report on the reproduction, 
and (b) the software used to obtain the results as a Docker container 
(controlling variation from dependencies and run-time environments) on GitLab. Submissions were reviewed in great detail, submitted code was test-run and checked for hard-coding of results.
11 out of 18 submissions were judged to conform with requirements. One original paper \cite{vajjala-rama-2018-experiments} attracted four reproductions \cite{bestgen-2020-reproducing,huber-coltekin-2020-reproduction,caines-buttery-2020-reprolang,arhiliuc-etal-2020-language} in what must be a groundbreaking first in NLP. See Table~\ref{tab:repro-attempts} for summaries of all 11 reproductions.
An aspect the organisers did not control was how to draw conclusions about reproducibility; most contributions provide binary conclusions but vary in how similar they require results to be for success.  
E.g.\ the four papers reproducing \citet{vajjala-rama-2018-experiments} all report similarly large deviations, but only one \cite{arhiliuc-etal-2020-language} concludes that the reproduction was not a success.

\section{Conclusions}\label{sec:conc}

It seemed so simple: share all data, code and parameter settings, and other researchers will be able to obtain the same results. Yet the systems we create remain stubbornly resistant to this goal: a tiny 
14.03\% of the 513 original/reproduction score pairs we looked at were the same. At the same time, worryingly small differences in code have been found to result in big differences in performance. 

Another striking finding is that reproduction under same conditions far more frequently yields results that are worse than results that are better: we found 258 out of 436 non-same reproduction results (59.2\%) to be worse, echoing findings from psychology \cite{open2015estimating}.  Why this should be the case for reproduction under \textit{same} conditions is unclear. It is probably to be expected for reproduction under \textit{different} conditions, as larger parameter spaces, more datasets and languages etc., are tested subsequently, and the original work may have selected better results.

There is a lot of work going on in NLP on reproducibility right now; it is to be hoped that we can solve the vexing and scientifically uninteresting problem of how to rerun code and get the same results
soon, and move on to addressing far more interesting questions of how reliable, stable and generalisable promising NLP results really are.

\section*{Acknowledgments}

We thank our reviewers for their valuable feedback. Shubham Agarwal's PhD fees are supported by Adeptmind Inc., Toronto, Canada. 

\bibliography{anthology,eacl2021}
\bibliographystyle{acl_natbib}

\newpage
\onecolumn
\section*{Appendices}
\vspace{.4cm}
\appendix

\section{Table of Reproductions Under Same Conditions}\label{app:table-same}

\begin{table*}[h!t]
    \centering
    \begin{small}
    \begin{tabular}{|p{3cm}|p{3cm}|p{3cm}|p{5.2cm}|}
    \hline
    Original paper & Reproduction study (same conditions) & NLP task & Summary of score differences \\
    \hline
    \citet{collins:m:1999} & \citet{gildea2001corpus} & Phrase-structure parsing    & +16.7\% error on Model 1 results \\
    \hline
    \citet{collins:m:1999} & \citet{bikel-2004-intricacies}& Phrase-structure parsing    & +11\% error on Model 2 results on WSJ00; later same results with help from Collins \\
    \hline
    \citet{freire2012approach} & \citet{van2013reusable} & NER & ``Despite feedback from Freire [...], results remained 20 points below those reported in Freire et al. (2012) in overall F-score" \cite{fokkens-etal-2013-offspring}\\
     \hline
    \citet{nakov-ng-2011-translating}& \citet{wieling-etal-2018-squib} & MT & *Unsuccessful (scripts did not work)\\
    \hline
    \citet{he-etal-2011-automatically} & \citet{wieling-etal-2018-squib} & Sentiment analysis & *-0.18 points \\
    \hline
    \citet{sauper-etal-2011-content} & \citet{wieling-etal-2018-squib} & Topic modelling & *Unsuccessful on 3 scores (8h cut-off reached)\\
    \hline
    \citet{liang-etal-2011-learning} & \citet{wieling-etal-2018-squib} & Question answering& *Exact reproduction of 2 scores in 4h\\
    \hline
    \citet{branavan-etal-2011-learning} & \citet{wieling-etal-2018-squib} & Joint learning of game strategy and text selection from game manual & *Unsuccessful on 7 scores (scripts did not generate output)\\
    \hline
    \citet{coavoux-crabbe-2016-neural} & \citet{wieling-etal-2018-squib} & Constituent parsing & *9/18 scores same, 9/18 parser did not complete for 4 languages\\ 
    \hline
    \citet{gao-etal-2016-physical} & \citet{wieling-etal-2018-squib} & Semantic role grounding & *Exact reproduction of 44/72 scores, 17 worse, 11 better, average -0.62 points\\
    \hline
    \citet{hu-etal-2016-harnessing} & \citet{wieling-etal-2018-squib} & Sentiment analysis, NER & *exact reproduction of 1/2 scores, 1 worse -0.2 points \\
    \hline
    \citet{nicolai-kondrak-2016-leveraging}& \citet{wieling-etal-2018-squib} & Stemming, lemmatisation & *2/8 scores -3.4 and -1.55 points, 6/8 scores took longer than 8h cut-off\\
    \hline
    \citet{tian-etal-2016-learning} & \citet{wieling-etal-2018-squib} & Sentence completion & *4/6 scores reproduced exactly, 2/6 differed -0.1 and +0.01 \%-points).\\
    \hline
    \citet{badjatiya2017deep} & \citet{fortuna-etal-2019-stop} & hate speech detection & reproduction under same conditions not possible due to issue with code; recreated/corrected system did well at OffensEval'19 but not at HatEval'19\\
    \hline
    \citet{choi-lee-2018-gist} & \citet{antonio-rodrigues-etal-2020-reproduction} & Argument Reasoning Comprehension Task & 1/1 score +0.002 points \\ 
    \hline
    \citet{zhao-etal-2018-blcu} & \citet{antonio-rodrigues-etal-2020-reproduction} & Argument Reasoning Comprehension Task & 1/1 score +0.038 points \\ 
    \hline
    \citet{tian-etal-2018-ecnu} & \citet{antonio-rodrigues-etal-2020-reproduction} & Argument Reasoning Comprehension Task & 1/1 score -0.021 points \\ 
        \hline
    \citet{niven-kao-2018-nlitrans} & \citet{antonio-rodrigues-etal-2020-reproduction} & Argument Reasoning Comprehension Task & 1/1 score +0.033  points \\ 
        \hline
    \citet{kim-etal-2018-snu} & \citet{antonio-rodrigues-etal-2020-reproduction} & Argument Reasoning Comprehension Task & 1/1 score -0.022 points \\ 
        \hline
    \citet{brassard-etal-2018-takelab} & \citet{antonio-rodrigues-etal-2020-reproduction} & Argument Reasoning Comprehension Task & Exact reproduction of 1/1 score.\\
    \hline
    \citet{artetxe-etal-2018-robust} & \citet{garneau-etal-2020-robust} & Cross-lingual Mappings of Word Embeddings & Main scores: 2/8 same, 1/8 -0.1, 5/8 +0.1 to +0.3; ablation scores: 4/40 scores same, 19/40 +0.1 to 6.9, 9/40 -0.1 to -0.9, 8/40 took too long \\ 
    \hline
    \citet{artetxe-etal-2018-robust} & \citet{plucinski-etal-2020-closer} & Cross-lingual Mappings of Word Embeddings & Main scores: 10/14 better, 4/14 worse; ablation scores: 3/48 scores same, 31/48 better, 14/48 worse \\ 
    \hline
    \citet{bohnet-etal-2018-morphosyntactic} & \citet{khoe-2020-reproducing} & POS and morphological tagging & POS tagging scores: 35/41 worse, 6/41 better; morph.\ tagging: 43/46 worse, 3/46 better \\ 
    \hline
    \citet{rotsztejn-etal-2018-eth} &
    \citet{rim-etal-2020-reproducing} & Relation extraction and classification (SemEval'18 T7) & 4 subtasks: 4/4 scores worse, up to 9.04 points; subtask 1.1 by relation: 3/6 worse, 3/6 better\\
    \hline
    \citet{nisioi-etal-2017-exploring} & \citet{cooper-shardlow-2020-combinmt} & Simplification & NTS default system: 1/2 automatic scores better, 1/2 automatic scores worse; 2/2 human scores worse \\
    \hline
    \multicolumn{4}{l}{}\\
    \multicolumn{4}{l}{\textit{Table continued on next page.}}
    \end{tabular}
    \end{small}
\end{table*}\begin{table*}[th!]
    \centering
    \begin{small}
\vspace{-7.25cm}    \begin{tabular}{|p{3cm}|p{3cm}|p{3cm}|p{5.2cm}|}
    \hline
    Original paper & Reproduction study (recreation of system) & NLP task & Summary of score differences \\
    \hline
    \citet{vajjala-rama-2018-experiments} & \citet{bestgen-2020-reproducing} & Automatic essay scoring (classification) &  multilingual: 6/11 better, 5/11 worse; monolingual: 15/27 better, 11/27 worse, 1/27 same; crosslingual: 5/8 better, 1/8 worse, 2/8 same\\
    \hline
    \citet{vajjala-rama-2018-experiments} & \citet{huber-coltekin-2020-reproduction} & Automatic essay scoring (classification) &  multilingual: 3/11 better, 8/11 worse; monolingual: 8/27 better, 19/27 worse; crosslingual: 6/8 better, 2/8 worse\\
    \hline
    \citet{vajjala-rama-2018-experiments} & \citet{caines-buttery-2020-reprolang} & Automatic essay scoring (classification) &  multilingual: 9/11 better, 2/11 worse; monolingual: 14/27 better, 11/27 worse, 2/27 same; crosslingual: 1/8 better, 7/8 worse\\
    \hline
    \citet{vajjala-rama-2018-experiments} & \citet{arhiliuc-etal-2020-language} & Automatic essay scoring (classification) & multilingual: 11/11 worse; monolingual: 7/27 better, 20/27 worse; crosslingual: 1/8 better, 5/8 worse, 2/8 same\\
    \hline
    \citet{magistry2018etiquetage} & \citet{millour-etal-2020-repliquer} & POS tagging for Alsatian & baseline: same (0.78 Acc); main: worse (Acc 0.87 vs.\ 0.91)\\
    \hline
    \citet{howard2018universal} & \citet{abdellatif-elgammal-2020-ulmfit} & Sentiment classification, question classification, topic classification & 3/6 better, 3/6 worse  \\
    \hline
    \citet{vo2015target} & \citet{moore-rayson-2018-bringing} & Target Dependent Sentiment analysis & 6/6 better
    \\        
    \hline
    \citet{wang2017tdparse} & \citet{moore-rayson-2018-bringing} & Target Dependent Sentiment analysis & 2/5 better, 3/5 worse 
    \\        
    \hline
    \citet{tang2016effective} & \citet{moore-rayson-2018-bringing} & Target Dependent Sentiment analysis & 3/3 worse 
    \\        
    \hline
    \end{tabular}
    \caption{Tabular overview of individual repeatability tests from 34 paper pairs, and a total of 549 score pairs. * = additional information obtained from hyperlinked material.\\ 
    \textcolor{white}{ind} Where scores obtained in a repeatability study (reproduction under same conditions) are worse than in the original work, this should \textit{not} be interpreted as casting the original work in a negative light. This is because it is normally not possible to create the exact same conditions in repeatability studies (and lower scores can result from such differences), and because the outcome from multiple repeatability studies may be very different.\\
    \textcolor{white}{ind} For a small number of papers, the score pairs included in this table are a subset of scores reported in the paper. More generally, the summary in the last column should not be interpreted as a summary of the whole paper and its findings.\\
    \textcolor{white}{ind} Our intention here is to summarise differences that have been reported in the literature, rather than draw conclusions about what may have caused the differences.\\
    }
    \label{tab:repro-attempts}
    \end{small}
\end{table*}

\clearpage
\pagebreak

\section{Table of Reproductions Under Varied Conditions}\label{app:results-varied}

\begin{table*}[h!]
    \centering
    \begin{small}
    \begin{tabular}{|p{2.6cm}|p{3.4cm}|p{3cm}|p{5.2cm}|}
    \hline
    Original paper & reproduction study (confirmation of finding) & NLP task & Summary of outcome (as interpreted by authors) \\
    \hline
    \citet{plank-etal-2016-multilingual} & \citet{horsmann-zesch-2017-lstms} & POS tagging & Confirmed for finer-grained tagsets \\
    \hline
    \citet{shen2018neural} & \citet{htut-etal-2018-grammar,htut-etal-2018-grammar-induction} & Grammar induction & Overall finding confirmed (that PRPN is a high performing grammar induction method) \\
    \hline
    \citet{magistry2018etiquetage} & \citet{millour-etal-2020-repliquer} & POS tagging & Not confirmed, reproduction results worse by $>$ 10 BLEU points \\
    \hline
    \citet{vajjala-rama-2018-experiments} & \citet{arhiliuc-etal-2020-language} 
     & Automatic essay scoring (classification) & Lower classification results on a corpus of Asian learners' English. \\
    \hline
    \citet{vajjala-rama-2018-experiments} & \citet{caines-buttery-2020-reprolang}
     & Automatic essay scoring (classification) & Lower classification results for English and Spanish CEFR datasets, and some adversarial data (e.g., scrambled English texts).  \\
    \hline
    \citet{vajjala-rama-2018-experiments} & 
    \citet{huber-coltekin-2020-reproduction}
     & Automatic essay scoring (classification) & Lower classification results for English Cambridge Learner Corpus.  \\
    \hline
    \citet{artetxe-etal-2018-robust} & \citet{garneau-etal-2020-robust} & Cross-lingual mappings of word embeddings & For other distant language pairs (from English to Estonian, Latvian, Finnish, Persian) the method did not converge or obtained lower scores. \\
    \hline
    \citet{artetxe-etal-2018-robust} & 
    \citet{plucinski-etal-2020-closer} & Cross-lingual mappings of word embeddings &  For other distant language pairs (from English to Czech, Polish) the method did not converge or obtained lower scores.\\
    \hline
    \citet{howard2018universal} & **\citet{abdellatif-elgammal-2020-ulmfit} & Sentiment classification, question classification, topic classification & Confirmed that transfer learning (pre-training) improves final classification accuracy.\\
    \hline
    \end{tabular}
    \caption{Tabular overview of individual studies to confirm a previous research finding. * = additional information obtained from hyperlinked material; ** = reproduction study had minor differences, e.g.\ hyperparameter tuning was omitted \cite{abdellatif-elgammal-2020-ulmfit}.\\
    \textcolor{white}{ind} The comments from the caption for Table~\ref{tab:repro-attempts} also apply here, but note that some differences between original and reproduction study are overt and intentional in the case of the papers in this table, whereas they are not intentional and often inadvertent in the case of the papers in Table~\ref{tab:repro-attempts}.}
    \label{tab:findings-attempts}
    \end{small}
\end{table*}

\vfill
\newpage

\section{Verbatim VIM and ACM Definitions}\label{sec:appendix-a}

\begin{table*}[h]
\begin{small}
    \centering
\noindent\begin{tabular}{|p{5cm}|p{10cm}|}
\hline
\textbf{2.1} (2.1) \textbf{measurement} & process  of  experimentally  obtaining  one  or  more  \textbf{quantity  values}  that  can  reasonably  be  attributed  to a \textbf{quantity}\\
\hline
\textbf{2.15} \textbf{measurement precision} (precision) & closeness  of  agreement  between  \textbf{indications}  or  \textbf{measured  quantity  values}  obtained  by  replicate  \textbf{measurements}  on  the  same  or  similar  objects  under specified conditions \\
\hline
\textbf{2.20} (3.6, Notes 1 and 2) \textbf{repeatability condition of measurement} (repeatability condition) & condition  of  \textbf{measurement},  out  of  a  set  of  conditions   that   includes   the   same   \textbf{measurement procedure},   same   operators,   same   \textbf{measuring system},   same   operating   conditions   and   same   location,  and  replicate  measurements  on  the  same  or similar objects over a short period of time \\
\hline
\textbf{2.21} (3.6) \textbf{measurement repeatability} (repeatability) & \textbf{measurement  precision}  under  a  set  of  \textbf{repeatability conditions of measurement} \\
\hline
\textbf{2.24} (3.7, Note 2) \textbf{reproducibility condition of measurement} (reproducibility condition) & condition  of  \textbf{measurement},  out  of  a  set  of  conditions  that  includes  different  locations,  operators,  \textbf{measuring  systems},  and  replicate  measurements  on the same or similar objects \\
\hline
\textbf{2.25} (3.7) \textbf{measurement reproducibility} (reproducibility) & \textbf{measurement   precision}   under   \textbf{reproducibility conditions of measurement} \\
\hline
\textbf{2.3} (2.6) \textbf{measurand} & \textbf{quantity} intended to be measured\\
\hline
\end{tabular}
\end{small}
    \caption{VIM definitions of repeatability and reproducibility \cite{jcgm2012international}.}
    \label{tab:vim}
\end{table*}

\begin{table*}[h]
    \centering
    \begin{small}
\noindent\begin{tabular}{|p{3.4cm}|p{11.6cm}|}
\hline
\textbf{Repeatability} (Same team, same experimental setup) & The measurement can be obtained with stated precision by the same team using the same measurement procedure, the same measuring system, under the same operating conditions, in the same location on multiple trials. For computational experiments, this means that a researcher can reliably repeat her own computation. \\
\hline
\textbf{Reproducibility} (Different team, same experimental setup)* &         The measurement can be obtained with stated precision by a different team using the same measurement procedure, the same measuring system, under the same operating conditions, in the same or a different location on multiple trials. For computational experiments, this means that an independent group can obtain the same result using the author’s own artifacts.\\
\hline
\textbf{Replicability} (Different team, different experimental setup)* &         The measurement can be obtained with stated precision by a different team, a different measuring system, in a different location on multiple trials. For computational experiments, this means that an independent group can obtain the same result using artifacts which they develop completely independently.\\
\hline
\multicolumn{2}{l}{}\\
\hline
\multicolumn{2}{|p{15cm}|}{\underline{Results Validated}: 
This badge is applied to papers in which the main results of the paper have been successfully obtained by a person or team other than the author. Two levels are distinguished:\vspace{.1cm}}\\
%\hline

\multicolumn{1}{|l}{\hspace{.2cm}\underline{Results Reproduced v1.1}}  &     The main results of the paper have been obtained in a subsequent study by a person or team other than the authors, using, in part, artifacts provided by the author.\vspace{.1cm}\\
%\hline
\multicolumn{1}{|l}{\hspace{.2cm}\underline{Results Replicated v1.1}} &      The main results of the paper have been independently obtained in a subsequent study by a person or team other than the authors, without the use of author-supplied artifacts.\vspace{.1cm}\\

\multicolumn{2}{|p{15cm}|}{In each case, exact replication or reproduction of results is not required, or even expected. Instead, the results must be in agreement to within a tolerance deemed acceptable for experiments of the given type. In particular, differences in the results should not change the main claims made in the paper.}\\
\hline

    \end{tabular}
    \end{small}
    \caption{ACM definitions (bold) and badges (underlined) \cite{acm2020artifact}.}
    \label{tab:acm}
\end{table*}

\clearpage
\newpage
\onecolumn

\end{document}